\documentclass{styles/svproc}

\usepackage{url}
\usepackage{cite}

\begin{document}
\mainmatter

\title{Intelligence Across Embodiments}

\titlerunning{Intelligence Across Embodiments}

\author{
Bo Ai\inst{1,2} \and
Henrik I. Christensen\inst{2} \and
Hao Su\inst{3}
}

\authorrunning{B. Ai, H. I. Christensen, H. Su}

% \institute{
% Stanford University, Stanford, CA, USA
% \and
% University of California San Diego, La Jolla, CA, USA
% \and
% Sudo AI GmbH, Zurich, Switzerland
% }
\institute{
Stanford University
\and
University of California San Diego
\and
Sudo AI GmbH
}

\maketitle

\begin{abstract}

Robotic embodiment encompasses the sensing, kinematics, dynamics, geometry, actuation, and control through which an agent physically interacts with the world. These properties vary across robots and change over time. We argue that general embodied intelligence requires learning that accumulates across these differences. Prevailing methods that engineer correspondences to bridge embodiment differences offer immediate practical gains, but their assumptions limit the scope of transfer in the long run. Instead, a more general approach should discover representations that support transfer to a larger range of embodiments as experience grows. We propose embodiment diversity as a promising axis of scaling, and identify broad learned priors as a complementary ingredient. We call for evaluations that better characterize embodiment gaps and transfer performance. More broadly, cross-embodiment learning connects the practical challenge of learning from heterogeneous robot experience with a broader scientific pursuit inspired by nature—physical intelligence that adapts and co-evolves with its embodiments to gain agency over its behavior and physical forms.


\end{abstract}

\section*{Experience Beyond One Embodiment}

% 1. From programmed robotics to robot learning
Robotics has long sought machines capable of autonomously performing physical tasks in diverse environments. Traditionally, perception, planning, and control have provided computational foundations for specifying behavior through models and algorithms~\cite{lavalle2006planning,thrun2005probabilistic,lynch2017modern}. Machine learning offers a complementary approach, acquiring sensorimotor mappings from data and enabling capabilities difficult to specify manually~\cite{kumar2021rma,kaufmann2023champion,ai2025review,levine2016end}. Extending these successes to broadly capable robots depends on the experience available for learning. A widely discussed obstacle is the scarcity of robot data~\cite{Amato2025data,Goldberg2025Good}.

% 2. Embodiment mediates the value of data
Learning systems generally improve from two sources of experience: data generated by other agents, often humans, and data generated through their own interactions with an environment. The former can bootstrap the latter. Language models, for example, pre-train on human-generated text and subsequently improve specific capabilities, such as programming, through reinforcement learning~\cite{anthropic2026fable51,openai2026gpt6astra}; AlphaGo combined learning from human games with self-play~\cite{silver2016mastering}. Robot learning can draw on both sources in principle, but its experience is mediated by embodiment—the configuration of sensing, morphology, kinematics, dynamics, actuation, and low-level control through which an agent interacts with the world. Embodiment determines what a robot can perceive and do, which strategies it can realize, and how its interactions are manifested as data. Unlike language models that see the same languages from different texts, robot policies see distinct sensorimotor ``vocabularies'' given by different embodiments. 

% 3. Learning must span populations and generations of embodiments
This heterogeneity is intrinsic to embodied agents. Different goals and environments demand different physical interactions, which favor different embodiments. Embodiment variation occurs along two axes: across populations and over time. Across populations, biological evolution has produced diverse embodiments adapted to different ecological niches and ways of life~\cite{losos1990ecomorphology,gupta2021embodied}. For robot design, factories, homes, disaster sites, and extraterrestrial environments impose different requirements on geometry, mobility, precision, strength, compliance, and sensing~\cite{kirschner2025categorizing}. Even within the same design, variation persists because of manufacturing tolerances, payloads, calibration, sensors, and controller configurations. Across time, actuators wear, sensors drift, controllers are recalibrated, and successive hardware generations are redesigned manually or computationally~\cite{zhao2020robogrammar,gupta2021embodied,ha2026transformertransformer}. Thus, a general learning system needs to span both populations and generations of embodiments. Intelligence is the persistent entity in this continual learning process that spans changing physical entities.

\section*{From Engineered Alignment to Learned Representations}

% 4. From engineered alignment to learned correspondence
A common strategy for learning and transfer across embodiments is to engineer correspondences. At the hardware level, standardizing gripper designs aligns handheld data collection devices with the target end effector~\cite{chi2024universal,xu2025dexumi,fang2025dexop}, and low-cost leader arms reproduce the target robot’s kinematics for joint-space teleoperation~\cite{zhao2023learning,fu2024mobile}. Anthropomorphic hands and humanoids exploit structural similarities with humans to facilitate motion retargeting~\cite{wang2024dexcap,he2025omnih2o,fu2025humanplus,junge2025adapt}. At the level of model representations, image goals and flow predictions describe desired visual or geometric changes that can be shared across embodiments~\cite{physicalintelligence2026pi07,li2026egowam,he2025scaling,xu2024flow}. Interaction intent and contact provide interfaces for relating physical interactions across embodiments~\cite{wang2026kite,piseno2026cloak,tian2025diffusion}, and end-effector actions abstract away joint-level differences~\cite{chi2024universal,zha2026lap,he2025scaling}. These approaches establish shared interfaces with human knowledge to facilitate learning and transfer. 

Many of these approaches rely on human-designed priors whose assumptions about tasks and environments can become long-term bottlenecks. An end-effector action space assumes that specifying end-effector motion is sufficient for task completion, which becomes limiting when tasks demand whole-body coordination or particular dynamic responses. Contact-based correspondence presumes that the target embodiment can reproduce the relevant contacts within its geometric, kinematic, and dynamic constraints. When these assumptions fail, successful execution may require a different interaction strategy that the prescribed interface cannot express. Humanoid designs draw on humans as an existence proof of broad physical capability, but the diversity of embodiments in nature suggests that the human form need not be best suited to every physical task and environment~\cite{losos1990ecomorphology,gupta2021embodied}. These human-designed priors enable short-term progress, but their assumptions can constrain the improvement that we would like to see from increasing experience and computation. 

The more fundamental question is whether we can build systems that \textit{discover} the general representations needed for transfer through learning. A recurring pattern in artificial intelligence research is that representations learned with increasing data and computation can surpass human-designed priors~\cite{sutton2019bitter}. Computer vision moved from hand-designed features toward representations learned by neural networks~\cite{lowe2004distinctive,krizhevsky2012imagenet}. Chess systems combined search with policies and value functions learned through self-play, reducing reliance on manually designed game features~\cite{silver2018alphazero}. Analogously, cross-embodiment learning may discover abstractions ranging from embodiment-agnostic understanding to embodiment-specific predictions.

Such abstractions might capture task semantics (the intended outcomes of a command), spatial relations (the relative positions and orientations of objects and the agent), physical interactions (the consequences of actions), and control (joint coordination that realizes desired behavior). These categories describe one possible organization, and prior work explicitly models some of them. Learned representations need not follow this hierarchy and may instead encode invariances or relations that vary systematically with embodiment, such as latent action representations that associate different motor commands across embodiments with similar physical effects. The goal is for accumulating experience to improve these representations, broaden transfer, and reduce adaptation costs on new embodiments. The scientific question is what ingredients can support this cumulative improvement.

% Intuitively, such abstractions may include task semantics (the intended outcomes of a task command), spatial relations (positions and orientations of objects relative to one another and to the agent), physical interactions (the consequences of actions on the environment), and control (motor actuation that coordinates joints to achieve desired task-space behavior), some of which prior work has attempted to explicitly model. A learned policy may also discover representations outside human-defined hierarchies, such as latent action representations that vary systematically with embodiment structure. 
% % These historical lessons suggest that the capacity to discover representations is more crucial in the long run than reliance on human-designed priors. 
% % We should enable learners to develop this capacity as they accumulate experience. 
% The desired outcome is that growing experience yields better representations and lowers the cost of adaptation to new embodiments. This could provide a path toward general embodied intelligence that transfers readily to new embodiments. The scientific question is what ingredients can support this cumulative improvement.

\section*{What Enables Cumulative Transfer?}

Broader pretraining can make subsequent experience more useful. Human demonstrations produce greater gains after broad robot pretraining~\cite{kareer2025emergence}, while scaling pretraining on human video improves performance after robot-specific fine-tuning~\cite{dyna2026dyna2}. These findings suggest that the value of new data depends partly on the learner's prior representations. Humans offer an intuitive analogy: a pilot's understanding of an aircraft helps connect simulated experience to the controls, dynamics, and consequences of physical flight. Broad learned priors may similarly help robots interpret experience acquired through other embodiments and ground it in their own sensorimotor interfaces.

Embodiment diversity is a plausible source of such priors and should be treated as a primary axis of scaling alongside tasks and environments. Broader task pretraining can improve adaptation to new tasks~\cite{Barreiros2026acareful}, while data collected across environments supports generalization beyond the settings in which it was collected~\cite{Khazatsky2024droid}. Experience across embodiments may similarly reveal common interaction structure and how successful behavior changes with physical capabilities. 
% Useful diversity depends on the interactions it exposes: different gripper geometries may require different contacts, while differences in actuation or controller response may require different timing and feedback. Tasks must expose these differences for the learner to benefit from them. 
Early results support this direction. Training across diverse end effectors enables adaptation to new hands and tools~\cite{generalist2026thousandhands}, while simulation-based training supports zero-shot transfer to unseen embodiments in locomotion and dexterous manipulation~\cite{ai2025towards,patel2024getzero}. In locomotion, increasing embodiment diversity has also produced broader generalization than increasing data on fixed embodiments~\cite{ai2025towards}. Establishing how far these gains extend requires systematic variation in embodiment alongside tasks and environments.

Obtaining such experience poses practical challenges. Collecting diverse real-world embodiment data requires constructing, maintaining, and calibrating different physical systems, as well as collecting expert demonstrations on each. Simulation offers a practical platform for synthesizing diverse embodiments, varying their properties systematically, and studying transfer at scale. Generating embodiments requires mechanically feasible morphologies with compatible dynamics and actuation properties, together with controller settings that produce a range of closed-loop responses. Covering the range of embodiments encountered in practice therefore requires an understanding of the dependencies among geometry, mass distribution, actuator capabilities, and control. Generating demonstrations presents another challenge: planning can be computationally expensive, while reinforcement learning may require substantial reward engineering. Furthermore, simulated experience must improve learning and performance on real robots. Bridging the simulation-to-reality gap remains an open challenge, but recent results with visuomotor policies trained on simulated data show signs of promise~\cite{deshpande2026molmobot}. 

Training policies across a much larger population of embodiments raises fundamental questions about model architecture. Existing datasets typically contain on the order of $10^2$ embodiments, but systematic variation could expand this number by orders of magnitude. How should a model learn jointly from distinct observation and action spaces? What information about embodiment should be supplied explicitly, and what can be inferred through interaction? Models that rely on visual observations to identify embodiment may struggle as diversity grows, since visually similar robots can differ in dynamics and controller response. This motivates architectures that incorporate explicit robot descriptions~\cite{bohlinger2024onepolicy,ai2025towards,patel2024getzero} or infer embodiment properties from interaction history through in-context learning~\cite{liu2025locoformer,li2026rapid}. Richer cross-embodiment datasets would enable systematic investigation of these designs.

\section*{How to Evaluate Transfer?}

Scientific progress in cross-embodiment learning requires clearer evaluation of transfer. Calling a target robot ``unseen'' says little about the difficulty of generalization. Studies should specify how it differs from the training population, the control interface used, and the assumptions made about the task. They should also report what embodiment information and interaction data are available at test time, distinguishing zero-shot generalization from few-shot adaptation.

Task and embodiment are distinct but connected axes of generalization. Experience on a robot performing other tasks may already reveal its sensing, dynamics, and control responses. A policy can combine this experience with task information from other robots to solve a held-out task--embodiment combination~\cite{physicalintelligence2026pi07}. Evaluation should distinguish this setting from transfer to an entirely unseen embodiment where no data is ever collected. Determining which ``calibration experiences'' on a new robot most improve performance across other tasks is itself an important research question.

Evaluation should also distinguish the behavioral changes required for transfer. Motion-level transfer preserves task-space motion or contact patterns by adapting joint-level execution. Strategy-level transfer requires a different way of achieving the same goal. For example, shaking may unfold a garment effectively with one embodiment but exceed another robot’s dynamic capabilities, making slower, sequential unfolding more appropriate. This requires a policy to relate the task’s intended outcome to the robot’s capabilities and select a feasible interaction strategy. Preserving a motion and discovering an alternative strategy therefore test different transfer capabilities.

Simulation offers a practical platform for these evaluations. Existing work demonstrates its value for scalable policy evaluation~\cite{li2025evaluating,zhang2025real}, and dedicated benchmarks have begun to assess generalization across morphologies~\cite{parakh2025anybody}. Procedural generation could extend these efforts to diverse, mechanically plausible embodiments with systematic variation in sensing, morphology, dynamics, actuation, and control. Controlled comparisons would help accurately characterize a policy’s transfer behaviors, isolating which differences a policy accommodates, where transfer fails, and how these limits change as training experience grows.

% Evaluation protocols must likewise distinguish zero-shot transfer from few-shot adaptation, and transfer to a held-out embodiment from transfer to a held-out task--embodiment combination. Observing a target robot on other tasks changes the problem: source-robot experience can indicate what interaction achieves the goal, while prior experience on the target robot can reveal how that interaction should be realized through its sensorimotor interface. Benchmarks that vary tasks and embodiments systematically are therefore needed to determine where transfer occurs, what information enables it, and how its efficiency changes as experience accumulates.

\section*{Towards Adaptive and Self-Evolving Robots}

Cross-embodiment learning and transfer create broad possibilities for robotics. In nature, physical capabilities develop through adaptation within a lifetime and evolution across generations. Learning that accumulates across changing embodiments could connect these processes in artificial systems, allowing intelligence to increasingly shape the physical means through which it acts.

One route toward this vision is computational robot design. Computational co-design already optimizes physical form and control together, from simulated evolution~\cite{sims1994evolving} and automatically designed and fabricated machines~\cite{lipson2000automatic} to joint optimization of morphology and policy~\cite{schaff2019jointly,chen2020hardware,ha2026transformertransformer,bohlinger2026shape}. A central bottleneck is that evaluating each candidate embodiment requires establishing what it can achieve under an appropriate controller. Efficient cross-embodiment transfer could reduce this cost, enabling broader searches over morphology, sensing, and actuation for more demanding tasks. Designs could then be evaluated both for their immediate performance and for how readily they support learning new capabilities.

Computational robot design could become a part of the co-evolution process of embodiment and policy. Better policies could make previously difficult designs viable, and better designs could make new behaviors easier to learn. Computational studies inspired by biological evolution already show that evolved morphologies can facilitate learning and adaptation to new tasks~\cite{gupta2021embodied}. Cross-embodiment transfer could carry learned capabilities across successive designs, allowing each generation to build on experience acquired by earlier ones. New embodiments would, in turn, expose the learner to interactions and capabilities unavailable on previous platforms. Policy learning and embodiment design could thereby improve together, producing forms and behaviors better suited to the tasks and environments we care about.

Within an agent’s lifetime, embodiment could also become a means of adaptation. Biology offers a concrete precedent in red knots, which reversibly change their gizzard size in response to changes in diet~\cite{dekinga2001gizzard}. Artificial agents could adjust their physical configurations to changing demands by selecting tools, repositioning sensors, or reconfiguring limbs to gain reach, stability, or access. Cross-embodiment control could help them exploit these changes without learning each configuration from scratch. Selecting and modifying an embodiment would then become part of how an agent pursues its goals.

The practical challenge of reusing experience across robots thus leads to a broader scientific pursuit. Cross-embodiment learning could provide a foundation for physical intelligence that persists as its embodiments change and gains agency over those changes. Nature offers a precedent for the diversity of capabilities that adaptation and evolution can produce. The ambition of intelligence across embodiments is to build artificial agents that continually expand both what they can physically achieve and the physical means through which they can accomplish goals.

\section*{Acknowledgements}
We thank Zhanxin Wu, Marcel Torne, and Nico Bohlinger for helpful feedback.

\bibliography{references}

@article{ai2025review,
  author = {Bo Ai and Stephen Tian and Haochen Shi and Yixuan Wang and Tobias Pfaff and Cheston Tan and others},
  title = {{{A Review of Learning-Based Dynamics Models for Robotic Manipulation}}},
  journal = {Science Robotics},
  volume = {10},
  number = {106},
  pages = {eadt1497},
  year = {2025},
  doi = {10.1126/scirobotics.adt1497},
}

@inproceedings{tian2025diffusion,
  author = {Tongxuan Tian and Haoyang Li and Bo Ai and Xiaodi Yuan and Zhiao Huang and Hao Su},
  title = {{{Diffusion Dynamics Models with Generative State Estimation for Cloth Manipulation}}},
  booktitle = {Conference on Robot Learning (CoRL)},
  year = {2025},
}

@article{Amato2025data,
  author = {Amato, Nancy M. and Hutchinson, Seth and Garg, Animesh and Billard, Aude and Rus, Daniela and Tedrake, Russ and others},
  title = {{``Data will solve robotics and automation: True or false?'': A debate}},
  journal = {Science Robotics},
  volume = {10},
  number = {105},
  pages = {eaea7897},
  year = {2025},
  doi = {10.1126/scirobotics.aea7897},
  url = {https://www.science.org/doi/10.1126/scirobotics.aea7897},
}

@article{silver2016mastering,
  author = {Silver, David and Huang, Aja and Maddison, Chris J and Guez, Arthur and Sifre, Laurent and Van Den Driessche, George and others},
  title = {{Mastering the game of Go with deep neural networks and tree search}},
  journal = {nature},
  volume = {529},
  number = {7587},
  pages = {484--489},
  publisher = {Nature Publishing Group UK London},
  year = {2016},
  doi = {10.1038/nature16961},
}

@misc{openai2026gpt6astra,
  author = {{OpenAI}},
  title = {{GPT-6 Astra: A New Generation of Intelligence}},
  year = {2026},
  month = sep,
  howpublished = {OpenAI model release},
  url = {https://openai.com/index/gpt-6-astra/},
  urldate = {2026-09-06},
}

@misc{anthropic2026fable51,
  author = {{Anthropic}},
  title = {{Claude Fable 5.1}},
  year = {2026},
  month = sep,
  howpublished = {Anthropic model release},
  url = {https://www.anthropic.com/claude/fable},
  urldate = {2026-09-06},
}

@inproceedings{ai2025towards,
  author = {Bo Ai and Liu Dai and Nico Bohlinger and Dichen Li and Tongzhou Mu and Zhanxin Wu and others},
  title = {{{Towards Embodiment Scaling Laws in Robot Locomotion}}},
  booktitle = {Conference on Robot Learning (CoRL)},
  year = {2025},
}

@article{gupta2021embodied,
  author = {Gupta, Agrim and Savarese, Silvio and Ganguli, Surya and Fei-Fei, Li},
  title = {{Embodied intelligence via learning and evolution}},
  journal = {Nature Communications},
  volume = {12},
  number = {1},
  pages = {5721},
  year = {2021},
  doi = {10.1038/s41467-021-25874-z},
  url = {https://www.nature.com/articles/s41467-021-25874-z},
}

@article{he2025scaling,
  author = {Zihao He and Bo Ai and Tongzhou Mu and Yulin Liu and Weikang Wan and Jiawei Fu and others},
  title = {{{Scaling Cross-Embodiment World Models for Dexterous Manipulation}}},
  journal = {arXiv},
  year = {2025},
}

@article{dyna2026dyna2,
  author = {{Dyna Robotics}},
  title  = {Dyna-2: A 1-Million-Hour Scaling Law
for World-Action Models},
  year   = {2026},
  month  = {August},
  url    = {https://dyna.co/dyna-2},
}

@article{li2026egowam,
  title   = {EgoWAM: World Action Models Beyond Pixels with In-the-Wild Egocentric Human Data},
  author  = {Li, Baoyu and Yin, Xinchen and Lin, Mengying and Zhang, Yixin and Xu, Danfei},
  journal = {arXiv preprint arXiv:2607.08436},
  year    = {2026}
}

@inproceedings{chi2024universal,
  author = {Cheng Chi and Zhenjia Xu and Chuer Pan and Eric Cousineau and Benjamin Burchfiel and Siyuan Feng and others},
  editor = {Dana Kulic and Gentiane Venture and Kostas E. Bekris and Enrique Coronado},
  title = {{Universal Manipulation Interface: In-The-Wild Robot Teaching Without In-The-Wild Robots}},
  booktitle = {Robotics: Science and Systems XX, Delft, The Netherlands, July 15-19, 2024},
  year = {2024},
  doi = {10.15607/RSS.2024.XX.045},
  url = {https://doi.org/10.15607/RSS.2024.XX.045},
}

@misc{kareer2025emergence,
  author = {Kareer, Simar and Pertsch, Karl and Darpinian, James and Hoffman, Judy and Xu, Danfei and Levine, Sergey and others},
  title = {{Emergence of Human to Robot Transfer in Vision-Language-Action Models}},
  year = {2025},
  eprint = {2512.22414},
  archiveprefix = {arXiv},
  primaryclass = {cs.RO},
  url = {https://arxiv.org/abs/2512.22414},
}

@inproceedings{Khazatsky2024droid,
  author = {Alexander Khazatsky and Karl Pertsch and Suraj Nair and Ashwin Balakrishna and Sudeep Dasari and Siddharth Karamcheti and others},
  editor = {Dana Kulic and Gentiane Venture and Kostas E. Bekris and Enrique Coronado},
  title = {{{DROID:} {A} Large-Scale In-The-Wild Robot Manipulation Dataset}},
  booktitle = {Robotics: Science and Systems XX, Delft, The Netherlands, July 15-19, 2024},
  year = {2024},
  doi = {10.15607/RSS.2024.XX.120},
  url = {https://doi.org/10.15607/RSS.2024.XX.120},
}

@article{levine2016end,
  author = {Levine, Sergey and Finn, Chelsea and Darrell, Trevor and Abbeel, Pieter},
  title = {{End-to-End Training of Deep Visuomotor Policies}},
  journal = {Journal of Machine Learning Research},
  volume = {17},
  number = {39},
  pages = {1--40},
  year = {2016},
  url = {https://www.jmlr.org/papers/v17/15-522.html},
}

@misc{generalist2026thousandhands,
  author = {{Generalist Team}},
  title = {{Towards Machines with a Thousand Hands}},
  year = {2026},
  date = {2026-07-23},
  howpublished = {Generalist AI research blog},
  url = {https://generalistai.com/blog/towards-machines-with-a-thousand-hands},
}

@article{Barreiros2026acareful,
  author = {Jose Barreiros and Andrew Beaulieu and Aditya Bhat and Rick Cory and Eric Cousineau and Hongkai Dai and others},
  title = {{A careful examination of large behavior models for multitask dexterous manipulation}},
  journal = {Science Robotics},
  volume = {11},
  number = {113},
  pages = {eaea6201},
  year = {2026},
  doi = {10.1126/scirobotics.aea6201},
  url = {https://www.science.org/doi/abs/10.1126/scirobotics.aea6201},
}

@article{li2026rapid,
  title={Rapid Embodiment Adaptation for Quadrupedal Locomotion},
  author={Li, Dichen and Ai, Bo and Bohlinger, Nico and Peters, Jan and Su, Hao and Christensen, Henrik I},
  journal={arXiv preprint arXiv:2608.01506},
  year={2026}
}

@inproceedings{li2025evaluating,
  title     = {Evaluating Real-World Robot Manipulation Policies in Simulation},
  author    = {Li, Xuanlin and Hsu, Kyle and Gu, Jiayuan and
               Mees, Oier and Pertsch, Karl and Walke, Homer Rich and others},
  booktitle = {Proceedings of The 8th Conference on Robot Learning},
  series    = {Proceedings of Machine Learning Research},
  volume    = {270},
  pages     = {3705--3728},
  publisher = {PMLR},
  year      = {2025},
  url       = {https://proceedings.mlr.press/v270/li25c.html}
}

@article{zhang2025real,
  title   = {Real-to-Sim Robot Policy Evaluation with Gaussian Splatting
             Simulation of Soft-Body Interactions},
  author  = {Zhang, Kaifeng and Sha, Shuo and Jiang, Hanxiao and
             Loper, Matthew and Song, Hyunjong and Cai, Guangyan and others},
  journal = {arXiv preprint arXiv:2511.04665},
  year    = {2025},
  url     = {https://arxiv.org/abs/2511.04665}
}

@article{parakh2025anybody,
  title   = {{AnyBody}: A Benchmark Suite for Cross-Embodiment Manipulation},
  author  = {Parakh, Meenal and Kirchmeyer, Alexandre and
             Han, Beining and Deng, Jia},
  journal = {arXiv preprint arXiv:2505.14986},
  year    = {2025},
  url     = {https://arxiv.org/abs/2505.14986}
}

@misc{liu2025locoformer,
  author = {Liu, Min and Pathak, Deepak and Agarwal, Ananye},
  title = {{LocoFormer: Generalist Locomotion via Long-context Adaptation}},
  year = {2025},
  eprint = {2509.23745},
  archiveprefix = {arXiv},
  primaryclass = {cs.RO},
  url = {https://arxiv.org/abs/2509.23745},
}

@article{kumar2021rma,
  title={{RMA}: Rapid motor adaptation for legged robots},
  author={Kumar, Ashish and Fu, Zipeng and Pathak, Deepak and Malik, Jitendra},
  journal={arXiv preprint arXiv:2107.04034},
  year={2021}
}

@article{kaufmann2023champion,
  title={Champion-level drone racing using deep reinforcement learning},
  author={Kaufmann, Elia and Bauersfeld, Leonard and Loquercio, Antonio and M{\"u}ller, Matthias and Koltun, Vladlen and Scaramuzza, Davide},
  journal={Nature},
  volume={620},
  number={7976},
  pages={982--987},
  year={2023},
  publisher={Nature Publishing Group UK London}
}

@article{kirschner2025categorizing,
  title={Categorizing robots by performance fitness into the tree of robots},
  author={Kirschner, Robin Jeanne and Karacan, K{\"u}bra and Melone, Alessandro and Haddadin, Sami},
  journal={Nature Machine Intelligence},
  volume={7},
  number={3},
  pages={459--470},
  year={2025},
  publisher={Nature Publishing Group UK London}
}

@article{zhao2020robogrammar,
  author = {Zhao, Allan and Xu, Jie and Konakovi{\'c}-Lukovi{\'c}, Mina and Hughes, Josephine and Spielberg, Andrew and Rus, Daniela and Matusik, Wojciech},
  title = {{RoboGrammar: Graph Grammar for Terrain-Optimized Robot Design}},
  journal = {ACM Transactions on Graphics},
  volume = {39},
  number = {6},
  pages = {188:1--188:16},
  year = {2020},
  doi = {10.1145/3414685.3417831},
  url = {https://doi.org/10.1145/3414685.3417831},
}

@article{losos1990ecomorphology,
  author  = {Jonathan B. Losos},
  title   = {Ecomorphology, Performance Capability, and Scaling of
             West Indian Anolis Lizards: An Evolutionary Analysis},
  journal = {Ecological Monographs},
  volume  = {60},
  number  = {3},
  pages   = {369--388},
  year    = {1990},
  doi     = {10.2307/1943062}
}

@inproceedings{xu2024flow,
  author = {Xu, Mengda and Xu, Zhenjia and Xu, Yinghao and Chi, Cheng and Wetzstein, Gordon and Veloso, Manuela and others},
  title = {{Flow as the Cross-Domain Manipulation Interface}},
  booktitle = {Conference on Robot Learning},
  year = {2024},
}

@article{lipson2000automatic,
  author = {Lipson, Hod and Pollack, Jordan B.},
  title = {{Automatic Design and Manufacture of Robotic Lifeforms}},
  journal = {Nature},
  volume = {406},
  number = {6799},
  pages = {974--978},
  year = {2000},
  doi = {10.1038/35023115},
}

@inproceedings{patel2024getzero,
  author = {Austin Patel and Shuran Song},
  title = {{{GET-Zero}: Graph Embodiment Transformer for Zero-shot Embodiment Generalization}},
  booktitle = {2025 IEEE International Conference on Robotics and Automation (ICRA)},
  year = {2025},
}

@article{bohlinger2024onepolicy,
  author = {Bohlinger, Nico and Czechmanowski, Grzegorz and Krupka, Maciej and Kicki, Piotr and Walas, Krzysztof and Peters, Jan and others},
  title = {{One Policy to Run Them All: an End-to-end Learning Approach to Multi-Embodiment Locomotion}},
  journal = {Conference on Robot Learning},
  year = {2024},
}

@article{
Goldberg2025Good,
author = {Ken Goldberg },
title = {Good old-fashioned engineering can close the 100,000-year “data gap” in robotics},
journal = {Science Robotics},
volume = {10},
number = {105},
pages = {eaea7390},
year = {2025},
doi = {10.1126/scirobotics.aea7390},
URL = {https://www.science.org/doi/abs/10.1126/scirobotics.aea7390},
eprint = {https://www.science.org/doi/pdf/10.1126/scirobotics.aea7390}}

@inproceedings{schaff2019jointly,
  author = {Schaff, Charles and Yunis, David and Chakrabarti, Ayan and Walter, Matthew R},
  title = {{Jointly learning to construct and control agents using deep reinforcement learning}},
  booktitle = {2019 international conference on robotics and automation (ICRA)},
  pages = {9798--9805},
  organization = {IEEE},
  year = {2019},
}

@inproceedings{chen2020hardware,
  author = {Chen, Tianjian and He, Zhanpeng and Ciocarlie, Matei},
  title = {{Hardware as policy: Mechanical and computational co-optimization using deep reinforcement learning}},
  booktitle = {Proceedings of the 2020 Conference on Robot Learning},
  series = {Proceedings of Machine Learning Research},
  volume = {155},
  pages = {1158--1173},
  year = {2021},
  url = {https://proceedings.mlr.press/v155/chen21a.html},
}

@inproceedings{xu2025dexumi,
  author = {Xu, Mengda and Zhang, Han and Hou, Yifan and Xu, Zhenjia and Fan, Linxi and Veloso, Manuela and others},
  title = {{DexUMI: Using Human Hand as the Universal Manipulation Interface for Dexterous Manipulation}},
  booktitle = {Proceedings of The 9th Conference on Robot Learning},
  series = {Proceedings of Machine Learning Research},
  volume = {305},
  pages = {437--459},
  publisher = {PMLR},
  year = {2025},
  url = {https://proceedings.mlr.press/v305/xu25b.html},
}

@article{zhao2023learning,
  author = {Zhao, Tony Z and Kumar, Vikash and Levine, Sergey and Finn, Chelsea},
  title = {{Learning fine-grained bimanual manipulation with low-cost hardware}},
  journal = {Robotics: Science and Systems},
  year = {2023},
}

@article{wang2024dexcap,
  title = {DexCap: Scalable and Portable Mocap Data Collection System for Dexterous Manipulation},
  author = {Wang, Chen and Shi, Haochen and Wang, Weizhuo and Zhang, Ruohan and Fei-Fei, Li and Liu, C. Karen},
  journal = {arXiv preprint arXiv:2403.07788},
  year = {2024}
}

@article{physicalintelligence2026pi07,
  author = {{Physical Intelligence}},
  title = {{${\pi}_{0.7}$: a Steerable Generalist Robotic Foundation Model with Emergent Capabilities}},
  journal = {arXiv preprint arXiv:2604.15483},
  year = {2026},
}

@article{zha2026lap,
  author = {Zha, Lihan and Hancock, Asher J and Zhang, Mingtong and Yin, Tenny and Huang, Yixuan and Shah, Dhruv and others},
  title = {{LAP}: Language-action pre-training enables zero-shot cross-embodiment transfer},
  journal = {arXiv preprint arXiv:2602.10556},
  year = {2026},
}

@book{lynch2017modern,
  author = {Lynch, Kevin M. and Park, Frank C.},
  title = {{Modern Robotics: Mechanics, Planning, and Control}},
  publisher = {Cambridge University Press},
  year = {2017},
}

@book{thrun2005probabilistic,
  author = {Thrun, Sebastian and Burgard, Wolfram and Fox, Dieter},
  title = {{Probabilistic Robotics}},
  publisher = {MIT Press},
  year = {2005},
}

@book{lavalle2006planning,
  author = {LaValle, Steven M.},
  title = {{Planning Algorithms}},
  publisher = {Cambridge University Press},
  year = {2006},
  url = {https://lavalle.pl/planning/},
}

@inproceedings{krizhevsky2012imagenet,
  author = {Krizhevsky, Alex and Sutskever, Ilya and Hinton, Geoffrey E.},
  title = {{{ImageNet} Classification with Deep Convolutional Neural Networks}},
  booktitle = {Advances in Neural Information Processing Systems},
  volume = {25},
  year = {2012},
  url = {https://papers.nips.cc/paper_files/paper/2012/hash/c399862d3b9d6b76c8436e924a68c45b-Abstract.html},
}

@misc{sutton2019bitter,
  author = {Sutton, Richard S.},
  title = {{The Bitter Lesson}},
  year = {2019},
  howpublished = {Essay, March 13},
  url = {http://www.incompleteideas.net/IncIdeas/BitterLesson.html},
}

@article{piseno2026cloak,
  title={Cloak: Zero-Shot Cross-Embodiment Manipulation by Masking the End-Effector from the VLA},
  author={Piseno, Michael and Tevet, Guy and Liu, C Karen},
  journal={arXiv preprint arXiv:2606.22836},
  year={2026}
}

@misc{wang2026kite,
  author = {Wang, Qianxu and Fang, Kuan},
  title = {{{KITE}: Decoupling Kinematics and Interaction for Zero-Shot Cross-Embodiment Manipulation}},
  year = {2026},
  eprint = {2606.22113},
  archiveprefix = {arXiv},
  primaryclass = {cs.RO},
  url = {https://arxiv.org/abs/2606.22113},
}

@article{fu2024mobile,
  title={Mobile {ALOHA}: Learning bimanual mobile manipulation with low-cost whole-body teleoperation},
  author={Fu, Zipeng and Zhao, Tony Z and Finn, Chelsea},
  journal={arXiv preprint arXiv:2401.02117},
  year={2024}
}

@misc{deshpande2026molmobot,
  author = {Deshpande, Abhay and Guru, Maya and Hendrix, Rose and Jauhri, Snehal and Eftekhar, Ainaz and Tripathi, Rohun and others},
  title = {{{MolmoB0T}: Large-Scale Simulation Enables Zero-Shot Manipulation}},
  year = {2026},
  eprint = {2603.16861},
  archiveprefix = {arXiv},
  primaryclass = {cs.RO},
  url = {https://arxiv.org/abs/2603.16861},
}

@misc{ha2026transformertransformer,
  author = {Ha, Huy and Liu, C. Karen and Song, Shuran},
  title = {{{Transformer Transformer}: A Unified Model for Motion-Conditioned Robot Co-design}},
  year = {2026},
  eprint = {2607.25798},
  archiveprefix = {arXiv},
  primaryclass = {cs.RO},
  url = {https://arxiv.org/abs/2607.25798},
}

@inproceedings{he2025omnih2o,
  title     = {{OmniH2O}: Universal and Dexterous Human-to-Humanoid Whole-Body Teleoperation and Learning},
  author    = {He, Tairan and Luo, Zhengyi and He, Xialin and Xiao, Wenli and
               Zhang, Chong and Zhang, Weinan and others},
  booktitle = {Proceedings of the 8th Conference on Robot Learning},
  series    = {Proceedings of Machine Learning Research},
  volume    = {270},
  pages     = {1516--1540},
  year      = {2025}
}

@inproceedings{fu2025humanplus,
  title     = {{HumanPlus}: Humanoid Shadowing and Imitation from Humans},
  author    = {Fu, Zipeng and Zhao, Qingqing and Wu, Qi and
               Wetzstein, Gordon and Finn, Chelsea},
  booktitle = {Proceedings of the 8th Conference on Robot Learning},
  series    = {Proceedings of Machine Learning Research},
  volume    = {270},
  pages     = {2828--2844},
  year      = {2025}
}

@article{junge2025adapt,
  title   = {{ADAPT-Teleop}: Robotic Hand with Human Matched Embodiment Enables Dexterous Teleoperated Manipulation},
  author  = {Junge, Kai and Hughes, Josie},
  journal = {npj Robotics},
  volume  = {3},
  number  = {1},
  pages   = {31},
  year    = {2025},
  doi     = {10.1038/s44182-025-00034-3}
}

@article{fang2025dexop,
  title={{DEXOP}: A device for robotic transfer of dexterous human manipulation},
  author={Fang, Hao-Shu and Romero, Branden and Xie, Yichen and Hu, Arthur and Huang, Bo-Ruei and Alvarez, Juan and others},
  journal={arXiv preprint arXiv:2509.04441},
  year={2025}
}

@article{lowe2004distinctive,
  author  = {Lowe, David G.},
  title   = {Distinctive Image Features from Scale-Invariant Keypoints},
  journal = {International Journal of Computer Vision},
  volume  = {60},
  number  = {2},
  pages   = {91--110},
  year    = {2004},
  doi     = {10.1023/B:VISI.0000029664.99615.94}
}

@inproceedings{sims1994evolving,
  author    = {Sims, Karl},
  title     = {Evolving Virtual Creatures},
  booktitle = {Proceedings of the 21st Annual Conference on Computer Graphics and Interactive Techniques},
  pages     = {15--22},
  year      = {1994},
  doi       = {10.1145/192161.192167}
}

@article{silver2018alphazero,
  author  = {Silver, David and Hubert, Thomas and Schrittwieser, Julian and
             Antonoglou, Ioannis and Lai, Matthew and Guez, Arthur and others},
  title   = {A General Reinforcement Learning Algorithm that Masters Chess,
             Shogi, and Go through Self-Play},
  journal = {Science},
  volume  = {362},
  number  = {6419},
  pages   = {1140--1144},
  year    = {2018},
  doi     = {10.1126/science.aar6404}
}

@article{dekinga2001gizzard,
  author  = {Dekinga, Anne and Dietz, Maurine W. and Koolhaas, Anita
             and Piersma, Theunis},
  title   = {Time Course and Reversibility of Changes in the Gizzards of
             Red Knots Alternately Eating Hard and Soft Food},
  journal = {Journal of Experimental Biology},
  volume  = {204},
  pages   = {2167--2173},
  year    = {2001},
  doi     = {10.1242/jeb.204.12.2167}
}

@article{bohlinger2026shape,
  title={Shape Your Body: Value Gradients for Multi-Embodiment Robot Design},
  author={Bohlinger, Nico and Peters, Jan},
  journal={arXiv preprint arXiv:2606.00702},
  year={2026}
}
\bibliographystyle{styles/bibtex/spmpsci}

\end{document}